\documentclass[10pt,journal,cspaper,compsoc]{IEEEtran}

\usepackage{amsmath,amssymb}
\usepackage{algorithm}
\usepackage{algorithmic}
\usepackage{multirow}

\usepackage[aboveskip=8pt]{caption}

\usepackage[dvips]{graphicx}
\DeclareGraphicsExtensions{.pdf}
\usepackage[american]{babel}

\usepackage{tabularx}

\usepackage[bookmarks=false,colorlinks=true,linkcolor=black,citecolor=black,filecolor=black,urlcolor=black]{hyperref}
\usepackage{url}
\usepackage{cvpr}
\usepackage{multicol}

\usepackage{stfloats}

\newcommand{\ct}[1]{\fontsize{7pt}{1pt}\selectfont{#1}}

\newcolumntype{x}{>\small c}

\def\cls{\mathit{cls}}
\def\reg{\mathit{reg}}

\begin{document}

\title{Faster R-CNN: Towards Real-Time Object Detection with Region Proposal Networks}

\author{Shaoqing~Ren,
        Kaiming~He,
        Ross Girshick,
        and~Jian~Sun
\IEEEcompsocitemizethanks{
\IEEEcompsocthanksitem S. Ren is with University of Science and Technology of China, Hefei, China. This work was done when S. Ren was an intern at Microsoft Research. Email: sqren@mail.ustc.edu.cn
\IEEEcompsocthanksitem K.~He and J.~Sun are with Visual Computing Group, Microsoft Research. E-mail: \{kahe,jiansun\}@microsoft.com
\IEEEcompsocthanksitem R.~Girshick is with Facebook AI Research. The majority of this work was done when R. Girshick was with Microsoft Research. E-mail: rbg@fb.com}
}

\IEEEcompsoctitleabstractindextext{%
\begin{abstract}
State-of-the-art object detection networks depend on region proposal algorithms to hypothesize object locations.
Advances like SPPnet \cite{He2014} and Fast R-CNN \cite{Girshick2015a} have reduced the running time of these detection networks, exposing region proposal computation as a bottleneck.
In this work, we introduce a \emph{Region Proposal Network} (RPN) that shares full-image convolutional features with the detection network, thus enabling nearly cost-free region proposals. An RPN is a fully convolutional network that simultaneously predicts object bounds and objectness scores at each position. The RPN is trained end-to-end to generate high-quality region proposals, which are used by Fast R-CNN for detection.
We further merge RPN and Fast R-CNN into a single network by sharing their convolutional features---using the recently popular terminology of neural networks with ``attention'' mechanisms, the RPN component tells the unified network where to look.
For the very deep VGG-16 model \cite{Simonyan2015}, our detection system has a frame rate of 5fps (\emph{including all steps}) on a GPU, while achieving state-of-the-art object detection accuracy on PASCAL VOC 2007, 2012, and MS COCO datasets with only 300 proposals per image. In ILSVRC and COCO 2015 competitions, Faster R-CNN and RPN are the foundations of the 1st-place winning entries in several tracks. Code has been made publicly available.
\end{abstract}


}

\maketitle

\IEEEpeerreviewmaketitle

\section{Introduction}

Recent advances in object detection are driven by the success of region proposal methods (\eg, \cite{Uijlings2013}) and region-based convolutional neural networks (R-CNNs) \cite{Girshick2014}. Although region-based CNNs were computationally expensive as originally developed in \cite{Girshick2014}, their cost has been drastically reduced thanks to sharing convolutions across proposals \cite{He2014,Girshick2015a}. The latest incarnation, Fast R-CNN \cite{Girshick2015a}, achieves near real-time rates using very deep networks \cite{Simonyan2015}, \emph{when ignoring the time spent on region proposals}. Now, proposals are the test-time computational bottleneck in state-of-the-art detection systems.

Region proposal methods typically rely on inexpensive features and economical inference schemes.
Selective Search \cite{Uijlings2013}, one of the most popular methods, greedily merges superpixels based on engineered low-level features. Yet when compared to efficient detection networks \cite{Girshick2015a}, Selective Search is an order of magnitude slower, at 2 seconds per image in a CPU implementation.
EdgeBoxes \cite{Zitnick2014} currently provides the best tradeoff between proposal quality and speed, at 0.2 seconds per image. Nevertheless, the region proposal step still consumes as much running time as the detection network.

One may note that fast region-based CNNs take advantage of GPUs, while the region proposal methods used in research are implemented on the CPU, making such runtime comparisons inequitable. An obvious way to accelerate proposal computation is to re-implement it for the GPU. This may be an effective engineering solution, but re-implementation ignores the down-stream detection network and therefore misses important opportunities for sharing computation.

\begin{figure*}[t]
\centering
\includegraphics[width=1.0\linewidth]{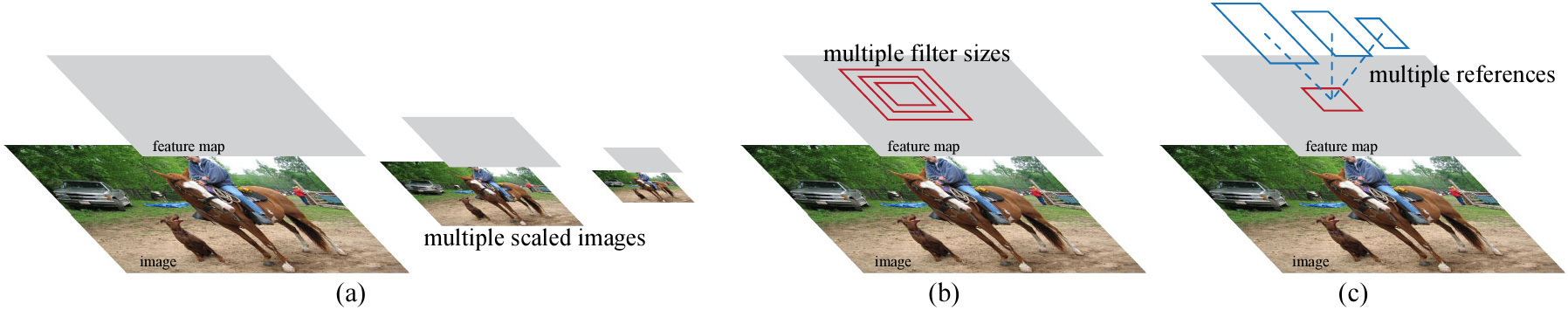}
\caption{Different schemes for addressing multiple scales and sizes. (a) Pyramids of images and feature maps are built, and the classifier is run at all scales. (b) Pyramids of filters with multiple scales/sizes are run on the feature map. (c) We use pyramids of reference boxes in the regression functions.}
\label{fig:pyramids}
\end{figure*}

In this paper, we show that an algorithmic change---computing proposals with a deep convolutional neural network---leads to an elegant and effective solution where proposal computation is nearly cost-free given the detection network's computation.
To this end, we introduce novel \emph{Region Proposal Networks} (RPNs) that share convolutional layers with state-of-the-art object detection networks \cite{He2014,Girshick2015a}. By sharing convolutions at test-time, the marginal cost for computing proposals is small (\eg, 10ms per image).

Our observation is that the convolutional feature maps used by region-based detectors, like Fast R-CNN, can also be used for generating region proposals.
On top of these convolutional features, we construct an RPN by adding a few additional convolutional layers that simultaneously regress region bounds and objectness scores at each location on a regular grid.
The RPN is thus a kind of fully convolutional network (FCN) \cite{Long2015} and can be trained end-to-end specifically for the task for generating detection proposals.

RPNs are designed to efficiently predict region proposals with a wide range of scales and aspect ratios. In contrast to prevalent methods \cite{Felzenszwalb2010,Sermanet2014,He2014,Girshick2015a} that use pyramids of images (Figure~\ref{fig:pyramids}, a) or pyramids of filters (Figure~\ref{fig:pyramids}, b), we introduce novel ``anchor'' boxes that serve as references at multiple scales and aspect ratios. Our scheme can be thought of as a pyramid of regression references (Figure~\ref{fig:pyramids}, c), which avoids enumerating images or filters of multiple scales or aspect ratios. This model performs well when trained and tested using single-scale images and thus benefits running speed.

To unify RPNs with Fast R-CNN \cite{Girshick2015a} object detection networks, we propose a training scheme that alternates between fine-tuning for the region proposal task and then fine-tuning for object detection, while keeping the proposals fixed.
This scheme converges quickly and produces a unified network with convolutional features that are shared between both tasks.\footnote{Since the publication of the conference version of this paper \cite{Ren2015a}, we have also found that RPNs can be trained jointly with Fast R-CNN networks leading to less training time.}

We comprehensively evaluate our method on the PASCAL VOC detection benchmarks \cite{Everingham2007} where RPNs with Fast R-CNNs produce detection accuracy better than the strong baseline of Selective Search with Fast R-CNNs. Meanwhile, our method waives nearly all computational burdens of Selective Search at test-time---the effective running time for proposals is just 10 milliseconds.
Using the expensive very deep models of \cite{Simonyan2015}, our detection method still has a frame rate of 5fps (\emph{including all steps}) on a GPU, and thus is a practical object detection system in terms of both speed and accuracy.
We also report results on the MS COCO dataset \cite{Lin2014} and investigate the improvements on PASCAL VOC using the COCO data.
Code has been made publicly available at \texttt{\url{https://github.com/shaoqingren/faster_rcnn}} (in MATLAB) and \texttt{\url{https://github.com/rbgirshick/py-faster-rcnn}} (in Python).

A preliminary version of this manuscript was published previously \cite{Ren2015a}. Since then, the frameworks of RPN and Faster R-CNN have been adopted and generalized to other methods, such as 3D object detection \cite{Song2015}, part-based detection \cite{Zhu2015}, instance segmentation \cite{Dai2015a}, and image captioning \cite{Johnson2015}. Our fast and effective object detection system has also been built in commercial systems such as at Pinterests \cite{Kislyuk2015}, with user engagement improvements reported.

In ILSVRC and COCO 2015 competitions, Faster R-CNN and RPN are the basis of several 1st-place entries \cite{He2015a} in the tracks of ImageNet detection, ImageNet localization, COCO detection, and COCO segmentation. RPNs completely learn to propose regions from data, and thus can easily benefit from deeper and more expressive features (such as the 101-layer residual nets adopted in \cite{He2015a}). Faster R-CNN and RPN are also used by several other leading entries in these competitions\footnote{\url{http://image-net.org/challenges/LSVRC/2015/results}}. These results suggest that our method is not only a cost-efficient solution for practical usage, but also an effective way of improving object detection accuracy.

\section{Related Work}

\noindent\textbf{Object Proposals.} There is a large literature on object proposal methods. Comprehensive surveys and comparisons of object proposal methods can be found in \cite{Hosang2014,Hosang2015,Chavali2015}. Widely used object proposal methods include those based on grouping super-pixels (\eg, Selective Search \cite{Uijlings2013}, CPMC \cite{Carreira2012}, MCG \cite{Arbelaez2014}) and those based on sliding windows (\eg, objectness in windows \cite{Alexe2012}, EdgeBoxes \cite{Zitnick2014}). Object proposal methods were adopted as external modules independent of the detectors (\eg, Selective Search \cite{Uijlings2013} object detectors, R-CNN \cite{Girshick2014}, and Fast R-CNN \cite{Girshick2015a}).

\vspace{.5em}
\noindent\textbf{Deep Networks for Object Detection.} The R-CNN method \cite{Girshick2014} trains CNNs end-to-end to classify the proposal regions into object categories or background. R-CNN mainly plays as a classifier, and it does not  predict object bounds (except for refining by bounding box regression).
Its accuracy depends on the performance of the region proposal module (see comparisons in \cite{Hosang2015}).
Several papers have proposed ways of using deep networks for predicting object bounding boxes \cite{Szegedy2013,Sermanet2014,Erhan2014,Szegedy2014a}.
In the OverFeat method \cite{Sermanet2014}, a fully-connected layer is trained to predict the box coordinates for the localization task that assumes a single object. The fully-connected layer is then turned into a convolutional layer for detecting multiple class-specific objects. The MultiBox methods \cite{Erhan2014,Szegedy2014a} generate region proposals from a network whose last fully-connected layer simultaneously predicts multiple class-agnostic boxes, generalizing the ``single-box'' fashion of OverFeat. These class-agnostic boxes are used as proposals for R-CNN \cite{Girshick2014}.
The MultiBox proposal network is applied on a single image crop or multiple large image crops (\eg, 224$\times$224), in contrast to our fully convolutional scheme. MultiBox does not share features between the proposal and detection networks.
We discuss OverFeat and MultiBox in more depth later in context with our method.
Concurrent with our work, the DeepMask method \cite{Pinheiro2015} is developed for learning segmentation proposals.

Shared computation of convolutions \cite{Sermanet2014,He2014,Dai2015,Long2015,Girshick2015a} has been attracting increasing attention for efficient, yet accurate, visual recognition. The OverFeat paper \cite{Sermanet2014} computes convolutional features from an image pyramid for classification, localization, and detection. Adaptively-sized pooling (SPP) \cite{He2014} on shared convolutional feature maps is developed for efficient region-based object detection \cite{He2014,Ren2015} and semantic segmentation \cite{Dai2015}. Fast R-CNN \cite{Girshick2015a} enables end-to-end detector training on shared convolutional features and shows compelling accuracy and speed.

\begin{figure}[t]
\centering
\includegraphics[width=0.95\linewidth]{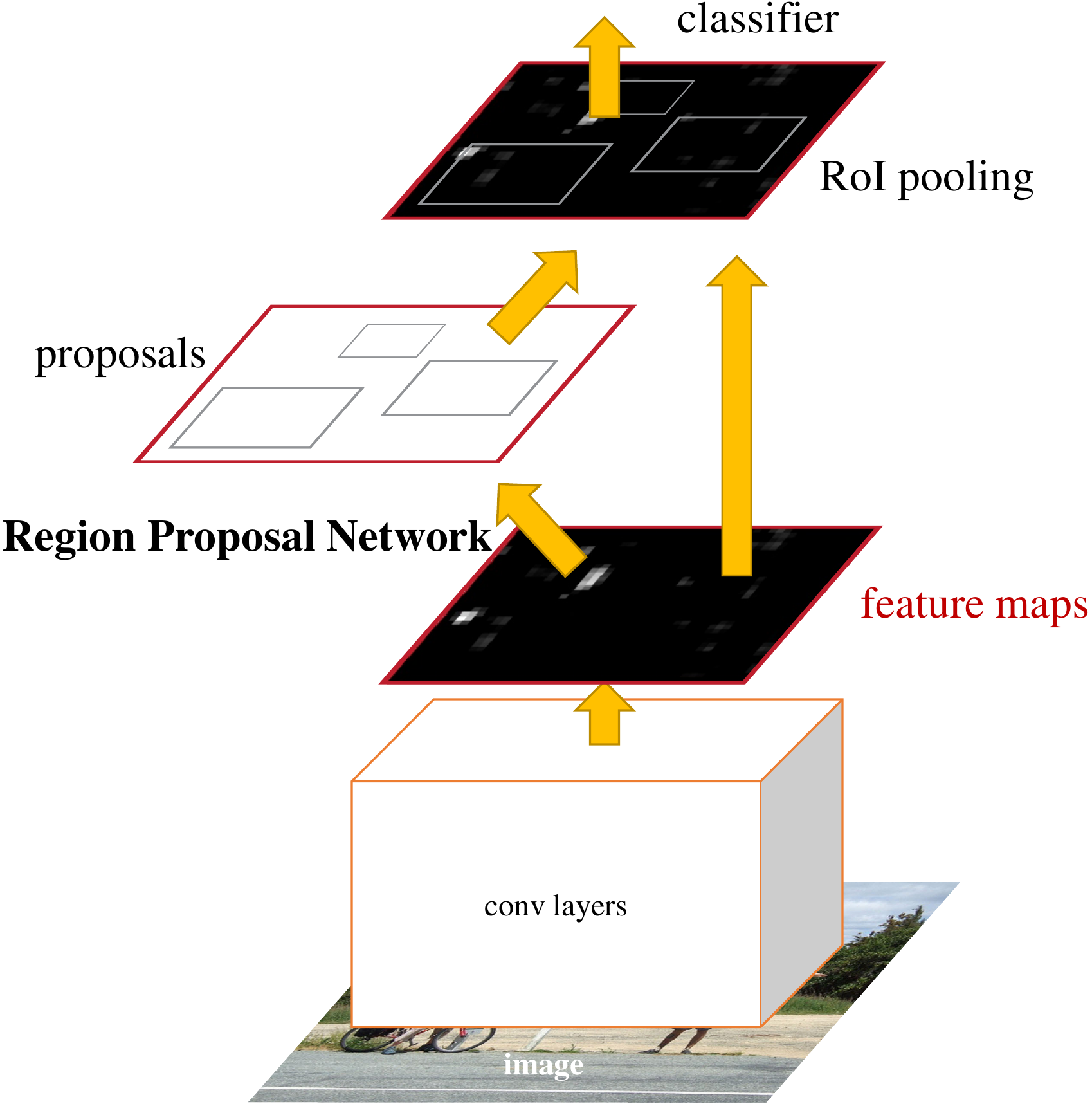}
\caption{Faster R-CNN is a single, unified network for object detection. The RPN module serves as the `attention' of this unified network.}
\label{fig:model}
\end{figure}

\begin{figure*}[t]
\centering
\includegraphics[width=0.98\linewidth]{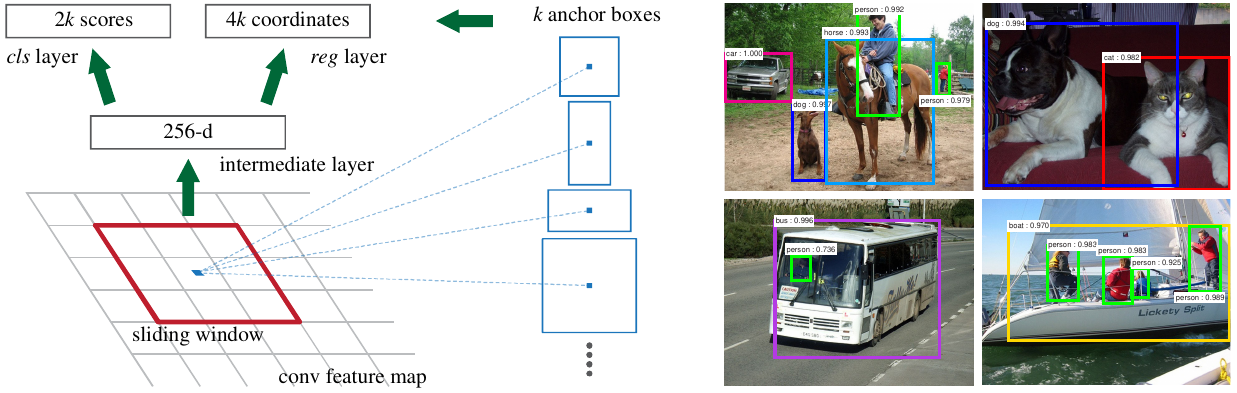}
\caption{\textbf{Left}: Region Proposal Network (RPN). \textbf{Right}: Example detections using RPN proposals on PASCAL VOC 2007 test. Our method detects objects in a wide range of scales and aspect ratios.}
\label{fig:rpn}
\end{figure*}

\section{Faster R-CNN}

Our object detection system, called Faster R-CNN, is composed of two modules. The first module is a deep fully convolutional network that proposes regions, and the second module is the Fast R-CNN detector \cite{Girshick2015a} that uses the proposed regions. The entire system is a single, unified network for object detection (Figure~\ref{fig:model}).
Using the recently popular terminology of neural networks with `attention' \cite{Chorowski2015} mechanisms, the RPN module tells the Fast R-CNN module where to look.
In Section~\ref{sec:rpn} we introduce the designs and properties of the network for region proposal. In Section~\ref{sec:training} we develop algorithms for training both modules with features shared.

\subsection{Region Proposal Networks}
\label{sec:rpn}

A Region Proposal Network (RPN) takes an image (of any size) as input and outputs a set of rectangular object proposals, each with an objectness score.\footnote{``Region'' is a generic term and in this paper we only consider \emph{rectangular} regions, as is common for many methods (\eg, \cite{Szegedy2014a,Uijlings2013,Zitnick2014}). ``Objectness'' measures membership to a set of object classes \vs~background.}
We model this process with a fully convolutional network \cite{Long2015}, which we describe in this section.
Because our ultimate goal is to share computation with a Fast R-CNN object detection network \cite{Girshick2015a}, we assume that both nets share a common set of convolutional layers.
In our experiments, we investigate the Zeiler and Fergus model \cite{Zeiler2014} (ZF), which has 5 shareable convolutional layers and the Simonyan and Zisserman model \cite{Simonyan2015} (VGG-16), which has 13 shareable convolutional layers.

To generate region proposals, we slide a small network over the convolutional feature map output by the last shared convolutional layer.
This small network takes as input an $n \times n$ spatial window of the input convolutional feature map.
Each sliding window is mapped to a lower-dimensional feature (256-d for ZF and 512-d for VGG, with ReLU \cite{Nair2010} following).
This feature is fed into two sibling fully-connected layers---a box-regression layer (\emph{reg}) and a box-classification layer (\emph{cls}).
We use $n=3$ in this paper, noting that the effective receptive field on the input image is large (171 and 228 pixels for ZF and VGG, respectively).
This mini-network is illustrated at a single position in Figure~\ref{fig:rpn} (left).
Note that because the mini-network operates in a sliding-window fashion, the fully-connected layers are shared across all spatial locations.
This architecture is naturally implemented with an $n \times n$ convolutional layer followed by two sibling $1 \times 1$ convolutional layers (for \emph{reg} and \emph{cls}, respectively). 


\subsubsection{Anchors}

At each sliding-window location, we simultaneously predict multiple region proposals, where the number of maximum possible proposals for each location is denoted as $k$.
So the \emph{reg} layer has $4k$ outputs encoding the coordinates of $k$ boxes, and the \emph{cls} layer outputs $2k$ scores that estimate probability of object or not object for each proposal\footnote{For simplicity we implement the \emph{cls} layer as a two-class softmax layer. Alternatively, one may use logistic regression to produce $k$ scores.}. The $k$ proposals are parameterized \emph{relative} to $k$ reference boxes, which we call \emph{anchors}. An anchor is centered at the sliding window in question, and is associated with a scale and aspect ratio (Figure~\ref{fig:rpn}, left). By default we use 3 scales and 3 aspect ratios, yielding $k=9$ anchors at each sliding position. For a convolutional feature map of a size $W \times H$ (typically $\sim$2,400), there are $WHk$ anchors in total.

\vspace{.5em}
\noindent\textbf{Translation-Invariant Anchors}

An important property of our approach is that it is \emph{translation invariant}, both in terms of the anchors and the functions that compute proposals relative to the anchors.
If one translates an object in an image, the proposal should translate and the same function should be able to predict the proposal in either location. This translation-invariant property is guaranteed by our method\footnote{As is the case of FCNs \cite{Long2015}, our network is translation invariant up to the network's total stride.}.
As a comparison, the MultiBox method \cite{Szegedy2014a} uses k-means to generate 800 anchors, which are \emph{not} translation invariant. So MultiBox does not guarantee that the same proposal is generated if an object is translated.

The translation-invariant property also reduces the model size.
MultiBox has a $(4+1)\times800$-dimensional fully-connected output layer, whereas our method has a $(4+2)\times9$-dimensional convolutional output layer in the case of $k=9$ anchors. As a result, our output layer has $2.8\times10^4$ parameters ($512\times(4+2)\times9$ for VGG-16), two orders of magnitude fewer than MultiBox's output layer that has $6.1\times10^6$ parameters ($1536\times(4+1)\times800$ for GoogleNet \cite{Szegedy2015} in MultiBox \cite{Szegedy2014a}).
If considering the feature projection layers, our proposal layers still have an order of magnitude fewer parameters than MultiBox\footnote{Considering the feature projection layers, our proposal layers' parameter count is $3\times3\times512\times512+512\times6\times9=2.4\times10^6$; MultiBox's proposal layers' parameter count is $7\times7\times(64+96+64+64)\times1536+1536\times5\times800=27\times10^6$.}. We expect our method to have less risk of overfitting on small datasets, like PASCAL VOC.

\vspace{.5em}
\noindent\textbf{Multi-Scale Anchors as Regression References}

Our design of anchors presents a novel scheme for addressing multiple scales (and aspect ratios).
As shown in Figure~\ref{fig:pyramids}, there have been two popular ways for multi-scale predictions. The first way is based on image/feature pyramids, \eg, in DPM \cite{Felzenszwalb2010} and CNN-based methods \cite{Sermanet2014,He2014,Girshick2015a}. The images are resized at multiple scales, and feature maps (HOG \cite{Felzenszwalb2010} or deep convolutional features \cite{Sermanet2014,He2014,Girshick2015a}) are computed for each scale (Figure~\ref{fig:pyramids}(a)). This way is often useful but is time-consuming. The second way is to use sliding windows of multiple scales (and/or aspect ratios) on the feature maps. For example, in DPM \cite{Felzenszwalb2010}, models of different aspect ratios are trained separately using different filter sizes (such as 5$\times$7 and 7$\times$5). If this way is used to address multiple scales, it can be thought of as a ``pyramid of filters'' (Figure~\ref{fig:pyramids}(b)). The second way is usually adopted jointly with the first way \cite{Felzenszwalb2010}.

As a comparison, our anchor-based method is built on \emph{a pyramid of anchors}, which is more cost-efficient. Our method classifies and regresses bounding boxes with reference to anchor boxes of multiple scales and aspect ratios.
It only relies on images and feature maps of a single scale, and uses filters (sliding windows on the feature map) of a single size. We show by experiments the effects of this scheme for addressing multiple scales and sizes (Table~\ref{tab:anchors}).

Because of this multi-scale design based on anchors, we can simply use the convolutional features computed on a single-scale image, as is also done by the Fast R-CNN detector \cite{Girshick2015a}. The design of multi-scale anchors is a key component for sharing features without extra cost for addressing scales.

\subsubsection{Loss Function}

For training RPNs, we assign a binary class label (of being an object or not) to each anchor.
We assign a positive label to two kinds of anchors: (i) the anchor/anchors with the highest Intersection-over-Union (IoU) overlap with a ground-truth box, \emph{or} (ii) an anchor that has an IoU overlap higher than 0.7 with any ground-truth box.
Note that a single ground-truth box may assign positive labels to multiple anchors. Usually the second condition is sufficient to determine the positive samples; but we still adopt the first condition for the reason that in some rare cases the second condition may find no positive sample.
We assign a negative label to a non-positive anchor if its IoU ratio is lower than 0.3 for all ground-truth boxes.
Anchors that are neither positive nor negative do not contribute to the training objective.

With these definitions, we minimize an objective function following the multi-task loss in Fast R-CNN \cite{Girshick2015a}. Our loss function for an image is defined as:
\begin{equation}\label{eq:loss}
\begin{aligned}
L(\{p_i\}, \{t_i\}) = \frac{1}{N_{\cls}}\sum_i L_{\cls}(p_i, p^{*}_i) \\ + \lambda\frac{1}{N_{\reg}}\sum_i  p^{*}_i L_{\reg}(t_i, t^{*}_i).
\end{aligned}
\end{equation}
Here, $i$ is the index of an anchor in a mini-batch and $p_i$ is the predicted probability of anchor $i$ being an object. The ground-truth label $p^{*}_i$ is 1 if the anchor is positive, and is 0 if the anchor is negative. $t_i$ is a vector representing the 4 parameterized coordinates of the predicted bounding box, and $t^{*}_i$ is that of the ground-truth box associated with a positive anchor.
The classification loss $L_{\cls}$ is log loss over two classes (object \vs~not object).
For the regression loss, we use $L_{\reg}(t_i, t^{*}_i)=R(t_i - t^{*}_i)$ where $R$ is the robust loss function (smooth L$_1$) defined in \cite{Girshick2015a}. The term $p^{*}_i L_{\reg}$ means the regression loss is activated only for positive anchors ($p^{*}_i=1$) and is disabled otherwise ($p^{*}_i=0$). The outputs of the \emph{cls} and \emph{reg} layers consist of $\{p_i\}$ and $\{t_i\}$ respectively.

The two terms are normalized by $N_{\cls}$ and $N_{\reg}$ and weighted by a balancing parameter $\lambda$. In our current implementation (as in the released code), the $cls$ term in Eqn.(\ref{eq:loss}) is normalized by the mini-batch size (\ie, $N_{\cls}=256$) and the $reg$ term is normalized by the number of anchor locations (\ie, $N_{\reg}\sim2,400$). By default we set $\lambda=10$, and thus both \emph{cls} and \emph{reg} terms are roughly equally weighted. We show by experiments that the results are insensitive to the values of $\lambda$ in a wide range (Table~\ref{tab:lambda}).
We also note that the normalization as above is not required and could be simplified.

For bounding box regression, we adopt the parameterizations of the 4 coordinates following \cite{Girshick2014}:
\begin{equation}
\begin{aligned}
t_{\textrm{x}} &=  (x - x_{\textrm{a}})/w_{\textrm{a}},\quad
t_{\textrm{y}} = (y - y_{\textrm{a}})/h_{\textrm{a}},\\
t_{\textrm{w}} &= \log(w / w_{\textrm{a}}), \quad
t_{\textrm{h}} = \log(h / h_{\textrm{a}}),\\
t^{*}_{\textrm{x}} &=  (x^{*} - x_{\textrm{a}})/w_{\textrm{a}},\quad
t^{*}_{\textrm{y}} = (y^{*} - y_{\textrm{a}})/h_{\textrm{a}},\\
t^{*}_{\textrm{w}} &= \log(w^{*} / w_{\textrm{a}}),\quad
t^{*}_{\textrm{h}} = \log(h^{*} / h_{\textrm{a}}),
\end{aligned}
\end{equation}
where $x$, $y$, $w$, and $h$ denote the box's center coordinates and its width and height.
Variables $x$, $x_{\textrm{a}}$, and $x^{*}$ are for the predicted box, anchor box, and ground-truth box respectively (likewise for $y, w, h$). This can be thought of as bounding-box regression from an anchor box to a nearby ground-truth box.

Nevertheless, our method achieves bounding-box regression by a different manner from previous RoI-based (Region of Interest) methods \cite{He2014,Girshick2015a}.
In \cite{He2014,Girshick2015a}, bounding-box regression is performed on features pooled from \emph{arbitrarily} sized RoIs, and the regression weights are \emph{shared} by all region sizes. In our formulation, the features used for regression are of the \emph{same} spatial size ($3 \times 3$) on the feature maps. To account for varying sizes, a set of $k$ bounding-box regressors are learned. Each regressor is responsible for one scale and one aspect ratio, and the $k$ regressors do \emph{not} share weights. As such, it is still possible to predict boxes of various sizes even though the features are of a fixed size/scale, thanks to the design of anchors.

\subsubsection{Training RPNs}
\label{sec:training_rpn}

The RPN can be trained end-to-end by back-propagation and stochastic gradient descent (SGD) \cite{LeCun1989}.
We follow the ``image-centric'' sampling strategy from \cite{Girshick2015a} to train this network. Each mini-batch arises from a single image that contains many positive and negative example anchors. It is possible to optimize for the loss functions of all anchors, but this will bias towards negative samples as they are dominate. Instead, we randomly sample 256 anchors in an image to compute the loss function of a mini-batch, where the sampled positive and negative anchors have a ratio of \emph{up to} 1:1. If there are fewer than 128 positive samples in an image, we pad the mini-batch with negative ones.

We randomly initialize all new layers by drawing weights from a zero-mean Gaussian distribution with standard deviation 0.01. All other layers (\ie, the shared convolutional layers) are initialized by pre-training a model for ImageNet classification \cite{Russakovsky2015}, as is standard practice \cite{Girshick2014}. We tune all layers of the ZF net, and conv3$\_1$ and up for the VGG net to conserve memory \cite{Girshick2015a}.
We use a learning rate of 0.001 for 60k mini-batches, and 0.0001 for the next 20k mini-batches on the PASCAL VOC dataset. We use a momentum of 0.9 and a weight decay of 0.0005 \cite{Krizhevsky2012}. Our implementation uses Caffe \cite{Jia2014}.

\subsection{Sharing Features for RPN and Fast R-CNN}
\label{sec:training}

Thus far we have described how to train a network for region proposal generation, without considering the region-based object detection CNN that will utilize these proposals.
For the detection network, we adopt Fast R-CNN \cite{Girshick2015a}.
Next we describe algorithms that learn a unified network composed of RPN and Fast R-CNN with shared convolutional layers (Figure~\ref{fig:model}).

\setlength{\tabcolsep}{3pt}
\renewcommand{\arraystretch}{1.1}
\begin{table*}[t]
\begin{center}
\caption{the learned average proposal size for each anchor using the ZF net (numbers for $s=600$).}
\vspace{-1em}
\small
\begin{tabular}{c||c|c|c||c|c|c||c|c|c}
  anchor & $128^2$, 2:1 & $128^2$, 1:1 & $128^2$, 1:2 & $256^2$, 2:1  & $256^2$, 1:1 & $256^2$, 1:2  & $512^2$, 2:1  & $512^2$, 1:1  & $512^2$, 1:2  \\
  \hline
  proposal & 188$\times$111 & 113$\times$114 & 70$\times$92 & 416$\times$229 & 261$\times$284 & 174$\times$332 & 768$\times$437 & 499$\times$501 & 355$\times$715  \\
\end{tabular}
\label{tab:anchorsZF}
\end{center}
\end{table*}

Both RPN and Fast R-CNN, trained independently, will modify their convolutional layers in different ways.
We therefore need to develop a technique that allows for sharing convolutional layers between the two networks, rather than learning two separate networks. We discuss three ways for training networks with features shared:

(i) \emph{Alternating training}. In this solution, we first train RPN, and use the proposals to train Fast R-CNN. The network tuned by Fast R-CNN is then used to initialize RPN, and this process is iterated. This is the solution that is used in all experiments in this paper.

(ii) \emph{Approximate joint training}. In this solution, the RPN and Fast R-CNN networks are merged into one network during training as in Figure~\ref{fig:model}. In each SGD iteration, the forward pass generates region proposals which are treated just like fixed, pre-computed proposals when training a Fast R-CNN detector.
The backward propagation takes place as usual, where for the shared layers the backward propagated signals from both the RPN loss and the Fast R-CNN loss are combined. This solution is easy to implement. But this solution ignores the derivative \wrt the proposal boxes' coordinates that are also network responses, so is approximate.
In our experiments, we have empirically found this solver produces close results, yet reduces the training time by about 25-50\% comparing with alternating training. This solver is included in our released Python code.

(iii) \emph{Non-approximate joint training}. As discussed above, the bounding boxes predicted by RPN are also functions of the input. The RoI pooling layer \cite{Girshick2015a} in Fast R-CNN accepts the convolutional features and also the predicted bounding boxes as input, so a theoretically valid backpropagation solver should also involve gradients \wrt the box coordinates. These gradients are ignored in the above approximate joint training.
In a non-approximate joint training solution, we need an RoI pooling layer that is differentiable \wrt the box coordinates.
This is a nontrivial problem and a solution can be given by an ``RoI warping'' layer as developed in \cite{Dai2015a}, which is beyond the scope of this paper.

\vspace{.5em}
\noindent\textbf{4-Step Alternating Training}. In this paper, we adopt a pragmatic 4-step training algorithm to learn shared features via alternating optimization.
In the first step, we train the RPN as described in Section~\ref{sec:training_rpn}. This network is initialized with an ImageNet-pre-trained model and fine-tuned end-to-end for the region proposal task.
In the second step, we train a separate detection network by Fast R-CNN using the proposals generated by the step-1 RPN. This detection network is also initialized by the ImageNet-pre-trained model. At this point the two networks do not share convolutional layers. In the third step, we use the detector network to initialize RPN training, but we fix the shared convolutional layers and only fine-tune the layers unique to RPN. Now the two networks share convolutional layers. Finally, keeping the shared convolutional layers fixed, we fine-tune the unique layers of Fast R-CNN. As such, both networks share the same convolutional layers and form a unified network. A similar alternating training can be run for more iterations, but we have observed negligible improvements.

\setlength{\tabcolsep}{8pt}
\renewcommand{\arraystretch}{1.1}
\begin{table*}[t]
\begin{center}
\caption{Detection results on \textbf{PASCAL VOC 2007 test set} (trained on VOC 2007 trainval). The detectors are Fast R-CNN with ZF, but using various proposal methods for training and testing.}
\vspace{-1em}
\small
\begin{tabular}{cc|cc|c}
  \multicolumn{2}{c|}{\bf train-time region proposals} & \multicolumn{2}{c|}{\bf test-time region proposals} &   \\
  method & \# boxes & method & \# proposals &  mAP (\%) \\
  \hline\hline
  SS & 2000 & SS & 2000 & 58.7 \\
  EB & 2000 & EB & 2000 & 58.6 \\
  RPN+ZF, shared & 2000 & RPN+ZF, shared & 300 & \textbf{59.9} \\
  \hline
  \hline
  \multicolumn{5}{l}{\emph{ablation experiments follow below}} \\
  \hline
  \hline
  RPN+ZF, unshared & 2000 & RPN+ZF, unshared & 300 & 58.7 \\
  \hline
  SS & 2000 & RPN+ZF & 100 & 55.1 \\
  SS & 2000 & RPN+ZF & 300 & 56.8 \\
  SS & 2000 & RPN+ZF & 1000 & 56.3 \\
  SS & 2000 & RPN+ZF (no NMS) & 6000 & 55.2 \\
  \hline
  SS & 2000 & RPN+ZF (no \emph{cls}) & 100 & 44.6 \\
  SS & 2000 & RPN+ZF (no \emph{cls}) & 300 & 51.4 \\
  SS & 2000 & RPN+ZF (no \emph{cls}) & 1000 & 55.8 \\
  \hline
  SS & 2000 & RPN+ZF (no \emph{reg}) & 300 & 52.1 \\
  SS & 2000 & RPN+ZF (no \emph{reg}) & 1000 & 51.3 \\
  \hline
  SS & 2000 & RPN+VGG & 300 & 59.2 \\
\end{tabular}
\label{tab:results}
\end{center}
\end{table*}

\subsection{Implementation Details}

We train and test both region proposal and object detection networks on images of a single scale \cite{He2014,Girshick2015a}. We re-scale the images such that their shorter side is $s=600$ pixels \cite{Girshick2015a}. Multi-scale feature extraction (using an image pyramid) may improve accuracy but does not exhibit a good speed-accuracy trade-off \cite{Girshick2015a}. On the re-scaled images, the total stride for both ZF and VGG nets on the last convolutional layer is 16 pixels, and thus is $\sim$10 pixels on a typical PASCAL image before resizing ($\sim$500$\times$375). Even such a large stride provides good results, though accuracy may be further improved with a smaller stride.

For anchors, we use 3 scales with box areas of $128^2$, $256^2$, and $512^2$ pixels, and 3 aspect ratios of 1:1,  1:2, and 2:1. These hyper-parameters are \emph{not} carefully chosen for a particular dataset, and we provide ablation experiments on their effects in the next section.
As discussed, our solution does not need an image pyramid or filter pyramid to predict regions of multiple scales, saving considerable running time.
Figure~\ref{fig:rpn} (right) shows the capability of our method for a wide range of scales and aspect ratios. Table~\ref{tab:anchorsZF} shows the learned average proposal size for each anchor using the ZF net.
We note that our algorithm allows predictions that are larger than the underlying receptive field. Such predictions are not impossible---one may still roughly infer the extent of an object if only the middle of the object is visible.

The anchor boxes that cross image boundaries need to be handled with care. During training, we ignore all cross-boundary anchors so they do not contribute to the loss. For a typical $1000\times600$ image, there will be roughly 20000 ($\approx60\times40\times9$) anchors in total. With the cross-boundary anchors ignored, there are about 6000 anchors per image for training. If the boundary-crossing outliers are not ignored in training, they introduce large, difficult to correct error terms in the objective, and training does not converge. During testing, however, we still apply the fully convolutional RPN to the entire image. This may generate cross-boundary proposal boxes, which we clip to the image boundary.

Some RPN proposals highly overlap with each other. To reduce redundancy, we adopt non-maximum suppression (NMS) on the proposal regions based on their \emph{cls} scores. We fix the IoU threshold for NMS at 0.7, which leaves us about 2000 proposal regions per image. As we will show, NMS does not harm the ultimate detection accuracy, but substantially reduces the number of proposals.
After NMS, we use the top-$N$ ranked proposal regions for detection. In the following, we train Fast R-CNN using 2000 RPN proposals, but evaluate different numbers of proposals at test-time.

\section{Experiments}

\subsection{Experiments on PASCAL VOC}

We comprehensively evaluate our method on the PASCAL VOC 2007 detection benchmark \cite{Everingham2007}. This dataset consists of about 5k trainval images and 5k test images over 20 object categories.
We also provide results on the PASCAL VOC 2012 benchmark for a few models.
For the ImageNet pre-trained network, we use the ``fast'' version of ZF net \cite{Zeiler2014} that has 5 convolutional layers and 3 fully-connected layers, and the public VGG-16 model\footnote{\url{www.robots.ox.ac.uk/~vgg/research/very_deep/}} \cite{Simonyan2015} that has 13 convolutional layers and 3 fully-connected layers.
We primarily evaluate detection mean Average Precision (mAP), because this is the actual metric for object detection (rather than focusing on object proposal proxy metrics).

Table~\ref{tab:results} (top) shows Fast R-CNN results when trained and tested using various region proposal methods. These results use the ZF net.
For Selective Search (SS) \cite{Uijlings2013}, we generate about 2000 proposals by the ``fast'' mode. For EdgeBoxes (EB) \cite{Zitnick2014}, we generate the proposals by the default EB setting tuned for 0.7 IoU.
SS has an mAP of 58.7\% and EB has an mAP of 58.6\% under the Fast R-CNN framework.
RPN with Fast R-CNN achieves competitive results, with an mAP of 59.9\% while using \emph{up to} 300 proposals\footnote{For RPN, the number of proposals (\eg, 300) is the maximum number for an image. RPN may produce fewer proposals after NMS, and thus the average number of proposals is smaller.}.
Using RPN yields a much faster detection system than using either SS or EB because of shared convolutional computations; the fewer proposals also reduce the region-wise fully-connected layers' cost (Table~\ref{tab:time}).

\setlength{\tabcolsep}{10pt}
\renewcommand{\arraystretch}{1.1}
\begin{table*}[t]
\begin{center}
\caption{Detection results on \textbf{PASCAL VOC 2007 test set}. The detector is Fast R-CNN and VGG-16. Training data: ``07'': VOC 2007 trainval, ``07+12'': union set of VOC 2007 trainval and VOC 2012 trainval. For RPN, the train-time proposals for Fast R-CNN are 2000. $^\dag$: this number was reported in \cite{Girshick2015a}; using the repository provided by this paper, this result is higher (68.1).}
\vspace{-1em}
\small
\begin{tabular}{cc|c|c}
  method & \# proposals & data & mAP (\%) \\
  \hline\hline
  SS & 2000 & 07 & 66.9$^\dag$ \\
  SS & 2000 & 07+12 & 70.0 \\
  \hline
  RPN+VGG, unshared & 300 & 07 & 68.5 \\
  RPN+VGG, shared & 300 & 07 & 69.9 \\
  RPN+VGG, shared & 300 & 07+12 & \textbf{73.2} \\
  \hline
  RPN+VGG, shared & 300 & COCO+07+12 & \textbf{78.8} \\
\end{tabular}
\label{tab:vgg}
\end{center}
\end{table*}

\begin{table*}[t]
\begin{center}
\caption{Detection results on \textbf{PASCAL VOC 2012 test set}. The detector is Fast R-CNN and VGG-16. Training data: ``07'': VOC 2007 trainval, ``07++12'': union set of VOC 2007 trainval+test and VOC 2012 trainval. For RPN, the train-time proposals for Fast R-CNN are 2000. \fontsize{8pt}{1em}\selectfont{$^\dag$: \url{http://host.robots.ox.ac.uk:8080/anonymous/HZJTQA.html}. $^\ddag$: \url{http://host.robots.ox.ac.uk:8080/anonymous/YNPLXB.html}. $^\S$: \url{http://host.robots.ox.ac.uk:8080/anonymous/XEDH10.html}.}}
\vspace{-1em}
\small
\begin{tabular}{cc|c|c}
  method & \# proposals & data & mAP (\%) \\
  \hline\hline
  SS & 2000 & 12 & 65.7 \\
  SS & 2000 & 07++12 & 68.4 \\
  \hline
  RPN+VGG, shared$^\dag$ & 300 & 12 & 67.0 \\
  RPN+VGG, shared$^\ddag$ & 300 & 07++12 & \textbf{70.4} \\
  \hline
  RPN+VGG, shared$^\S$ & 300 & COCO+07++12 & \textbf{75.9} \\
\end{tabular}
\label{tab:vgg12}
\end{center}
\end{table*}

\begin{table*}[t]
\begin{center}
\caption{\textbf{Timing} (ms) on a K40 GPU, except SS proposal is evaluated in a CPU. ``Region-wise'' includes NMS, pooling, fully-connected, and softmax layers. See our released code for the profiling of running time.}
\vspace{-1em}
\small
\begin{tabular}{c|c|ccc|c|c}
  model & system & conv & proposal & region-wise & total & rate \\
  \hline\hline
  VGG & SS + Fast R-CNN & 146 & 1510 & 174 & 1830 & 0.5 fps \\
  VGG & RPN + Fast R-CNN & 141 & \textbf{10} & 47 & \textbf{198} & \textbf{5 fps} \\
  \hline
  ZF & RPN + Fast R-CNN & 31 & \textbf{3} & 25 & \textbf{59}  & \textbf{17 fps}\\
\end{tabular}
\label{tab:time}
\end{center}
\end{table*}

\vspace{.5em}
\noindent\textbf{Ablation Experiments on RPN.}
To investigate the behavior of RPNs as a proposal method, we conducted several ablation studies.
First, we show the effect of sharing convolutional layers between the RPN and Fast R-CNN detection network.
To do this, we stop after the second step in the 4-step training process.
Using separate networks reduces the result slightly to 58.7\% (RPN+ZF, unshared, Table~\ref{tab:results}).
We observe that this is because in the third step when the detector-tuned features are used to fine-tune the RPN, the proposal quality is improved.

Next, we disentangle the RPN's influence on training the Fast R-CNN detection network.
For this purpose, we train a Fast R-CNN model by using the 2000 SS proposals and ZF net.
We fix this detector and evaluate the detection mAP by changing the proposal regions used at test-time.
In these ablation experiments, the RPN does not share features with the detector.

Replacing SS with 300 RPN proposals at test-time leads to an mAP of 56.8\%. The loss in mAP is because of the inconsistency between the training/testing proposals. This result serves as the baseline for the following comparisons.

Somewhat surprisingly, the RPN still leads to a competitive result (55.1\%) when using the top-ranked 100 proposals at test-time, indicating that the top-ranked RPN proposals are accurate. On the other extreme, using the top-ranked 6000 RPN proposals (without NMS) has a comparable mAP (55.2\%), suggesting NMS does not harm the detection mAP and may reduce false alarms.

Next, we separately investigate the roles of RPN's \emph{cls} and \emph{reg} outputs by turning off either of them at test-time.
When the \emph{cls} layer is removed at test-time (thus no NMS/ranking is used), we randomly sample $N$ proposals from the unscored regions. The mAP is nearly unchanged with $N=1000$ (55.8\%), but degrades considerably to 44.6\% when $N=100$. This shows that the \emph{cls} scores account for the accuracy of the highest ranked proposals.

On the other hand, when the \emph{reg} layer is removed at test-time (so the proposals become anchor boxes), the mAP drops to 52.1\%. This suggests that the high-quality proposals are mainly due to the regressed box bounds. The anchor boxes, though having multiple scales and aspect ratios, are not sufficient for accurate detection.

We also evaluate the effects of more powerful networks on the proposal quality of RPN alone. We use VGG-16 to train the RPN, and still use the above detector of SS+ZF. The mAP improves from 56.8\% (using RPN+ZF) to 59.2\% (using RPN+VGG).
This is a promising result, because it suggests that the proposal quality of RPN+VGG is better than that of RPN+ZF. Because proposals of RPN+ZF are competitive with SS (both are 58.7\% when consistently used for training and testing), we may expect RPN+VGG to be better than SS. The following experiments justify this hypothesis.

\newcolumntype{x}[1]{>{\centering}p{#1pt}}
\newcolumntype{y}{>{\centering}p{16pt}}
\newcommand{\hl}[1]{\textbf{\underline{#1}}}
\renewcommand{\arraystretch}{1.4}
\setlength{\tabcolsep}{4pt}
\begin{table*}[t]
\begin{center}
\small
\caption{Results on PASCAL VOC 2007 test set with Fast R-CNN detectors and VGG-16.
For RPN, the train-time proposals for Fast R-CNN are 2000. RPN$^*$ denotes the unsharing feature version. }
\vspace{-1em}
\label{tab:vgg_all}
\resizebox{\linewidth}{!}{
\begin{tabular}{x{24}|x{24}|x{54}|x{20}|yyyyyyyyyyyyyyyyyyyc}
  \ct{method} & \# box & data & mAP & \ct{areo} & \ct{bike} & \ct{bird} & \ct{boat} & \ct{bottle} & \ct{bus} & \ct{car} & \ct{cat} & \ct{chair} & \ct{cow} & \ct{table} & \ct{dog} & \ct{horse} & \ct{mbike} & \ct{person} & \ct{plant} & \ct{sheep} & \ct{sofa} & \ct{train} & \ct{tv} \\
  \hline\hline
  SS & 2000 & 07 & 66.9  & 74.5 & 78.3 & 69.2 & 53.2 & 36.6 & 77.3 & 78.2 & 82.0 & 40.7 & 72.7 & 67.9 & 79.6 & 79.2 & 73.0 & 69.0 & 30.1 & 65.4 & 70.2 & 75.8 & 65.8 \\
  SS & 2000 & 07+12 & 70.0  & {77.0} & 78.1 & 69.3 & 59.4 & 38.3 & 81.6 & 78.6 & {86.7} & 42.8 & 78.8 & {68.9} & 84.7 & 82.0 & 76.6 & 69.9 & 31.8 & 70.1 & {74.8} & 80.4 & 70.4 \\
    \hline
  RPN$^*$ & 300 & 07 & 68.5 & 74.1 & 77.2 & 67.7 & 53.9 & 51.0 & 75.1 & 79.2 & 78.9 & 50.7 & 78.0 & 61.1 & 79.1 & 81.9 & 72.2 & 75.9 & 37.2 & 71.4 & 62.5 & 77.4 & 66.4\\
  RPN & 300 & 07 & 69.9 & 70.0 & {80.6} & 70.1 & 57.3 & 49.9 & 78.2 & 80.4 & 82.0 & {52.2} & 75.3 & 67.2 & 80.3 & 79.8 & 75.0 & 76.3 & {39.1} & 68.3 & 67.3 & 81.1 & 67.6 \\
  RPN & 300 & 07+12 & {73.2} & 76.5 & 79.0 & {70.9} & {65.5} & {52.1} & {83.1} & {84.7} & 86.4 & 52.0 & {81.9} & 65.7 & {84.8} & {84.6} & {77.5} & {76.7} & 38.8 & {73.6} & 73.9 & {83.0} & {72.6}\\
  RPN & 300 & \footnotesize COCO+07+12 & \hl{78.8} & \hl{84.3} & \hl{82.0} & \hl{77.7} & \hl{68.9} & \hl{65.7} & \hl{88.1} & \hl{88.4} & \hl{88.9} & \hl{63.6} & \hl{86.3} & \hl{70.8} & \hl{85.9} & \hl{87.6} & \hl{80.1} & \hl{82.3} & \hl{53.6} & \hl{80.4} & \hl{75.8} & \hl{86.6} & \hl{78.9}
\end{tabular}
}
\end{center}
\end{table*}

\begin{table*}[t]
\begin{center}
\small
\caption{Results on PASCAL VOC 2012 test set with Fast R-CNN detectors and VGG-16.
For RPN, the train-time proposals for Fast R-CNN are 2000.}
\vspace{-1em}
\label{tab:vgg12_all}
\resizebox{\linewidth}{!}{
\begin{tabular}{x{24}|x{24}|x{54}|x{20}|yyyyyyyyyyyyyyyyyyyc}
   \ct{method} & \# box & data & mAP & \ct{areo} & \ct{bike} & \ct{bird} & \ct{boat} & \ct{bottle} & \ct{bus} & \ct{car} & \ct{cat} & \ct{chair} & \ct{cow} & \ct{table} & \ct{dog} & \ct{horse} & \ct{mbike} & \ct{person} & \ct{plant} & \ct{sheep} & \ct{sofa} & \ct{train} & \ct{tv} \\
  \hline\hline
  SS & 2000 & 12  & 65.7 & 80.3 & 74.7 & 66.9 & 46.9 & 37.7 & 73.9 & 68.6 & 87.7 & 41.7 & 71.1 & 51.1 & 86.0 & 77.8 & 79.8 & 69.8 & 32.1 & 65.5 & 63.8 & 76.4 & 61.7 \\
  SS & 2000 & 07++12 & 68.4 & 82.3 & 78.4 & 70.8 & 52.3 & 38.7 & {77.8} & 71.6 & {89.3} & 44.2 & 73.0 & 55.0 & \textbf{87.5} & 80.5 & 80.8 & 72.0 & 35.1 & 68.3 & \hl{65.7} & 80.4 & {64.2} \\
    \hline
  RPN & 300 & 12 & 67.0 & 82.3 & 76.4 & 71.0 & 48.4 & 45.2 & 72.1 & 72.3 & 87.3 & 42.2 & 73.7 & 50.0 & 86.8 & 78.7 & 78.4 & 77.4 & 34.5 & 70.1 & 57.1 & 77.1 & 58.9\\
  RPN & 300 & 07++12 & {70.4} & {84.9} & {79.8} & {74.3} & {53.9} & {49.8} & 77.5 & {75.9} & 88.5 & {45.6} & {77.1} & {55.3} & 86.9 & {81.7} & {80.9} & {79.6} & {40.1} & {72.6} & 60.9 & {81.2} & 61.5\\
  RPN & 300 & \footnotesize COCO+07++12 & \hl{75.9} & \hl{87.4} & \hl{83.6} & \hl{76.8} & \hl{62.9} & \hl{59.6} & \hl{81.9} & \hl{82.0} & \hl{91.3} & \hl{54.9} & \hl{82.6} & \hl{59.0} & \hl{89.0} & \hl{85.5} & \hl{84.7} & \hl{84.1} & \hl{52.2} & \hl{78.9} & {65.5} & \hl{85.4} & \hl{70.2}
\end{tabular}
}
\end{center}
\end{table*}

\begin{figure*}[t]
\centering
\includegraphics[width=0.68\linewidth]{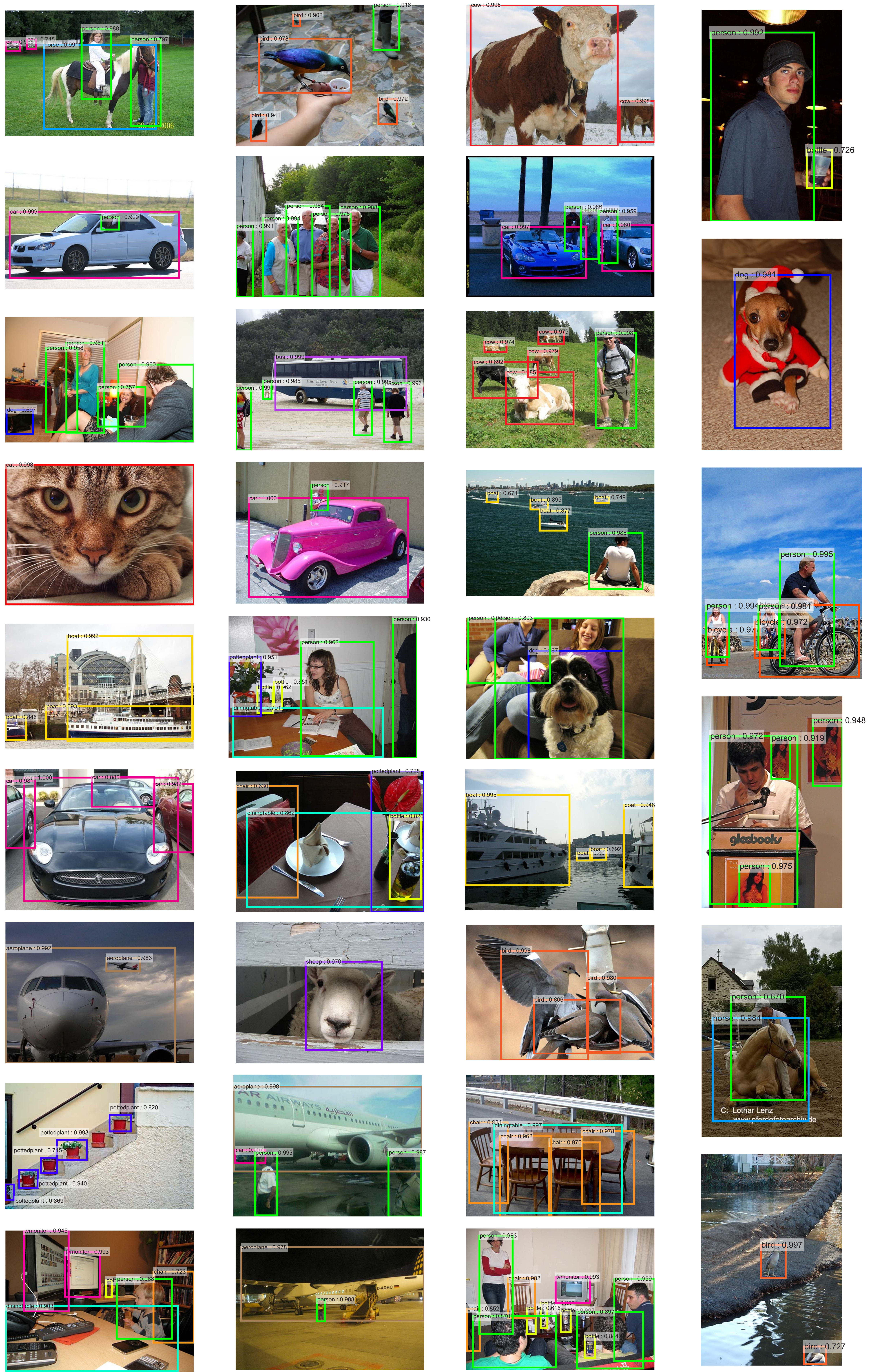}
\caption{Selected examples of object detection results on the PASCAL VOC 2007 test set using the Faster R-CNN system. The model is VGG-16 and the training data is 07+12 trainval (73.2\% mAP on the 2007 test set). Our method detects objects of a wide range of scales and aspect ratios. Each output box is associated with a category label and a softmax score in $[0,1]$. A score threshold of 0.6 is used to display these images. The running time for obtaining these results is \textbf{198ms} per image, \emph{including all steps}.}
\label{fig:results}
\end{figure*}

\vspace{.5em}
\noindent\textbf{Performance of VGG-16.}
Table~\ref{tab:vgg} shows the results of VGG-16 for both proposal and detection. Using RPN+VGG, the result is 68.5\% for \emph{unshared} features, slightly higher than the SS baseline. As shown above, this is because the proposals generated by RPN+VGG are more accurate than SS. Unlike SS that is pre-defined, the RPN is actively trained and benefits from better networks. For the feature-\emph{shared} variant, the result is 69.9\%---better than the strong SS baseline, yet with nearly cost-free proposals. We further train the RPN and detection network on the union set of PASCAL VOC 2007 trainval and 2012 trainval. The mAP is \textbf{73.2\%}.
Figure~\ref{fig:results} shows some results on the PASCAL VOC 2007 test set.
On the PASCAL VOC 2012 test set (Table~\ref{tab:vgg12}), our method has an mAP of \textbf{70.4\%} trained on the union set of VOC 2007 trainval+test and VOC 2012 trainval. Table~\ref{tab:vgg_all} and Table~\ref{tab:vgg12_all} show the detailed numbers.

In Table~\ref{tab:time} we summarize the running time of the entire object detection system. SS takes 1-2 seconds depending on content (on average about 1.5s), and Fast R-CNN with VGG-16 takes 320ms on 2000 SS proposals (or 223ms if using SVD on fully-connected layers \cite{Girshick2015a}). Our system with VGG-16 takes in total \textbf{198ms} for both proposal and detection. With the convolutional features shared, the RPN alone only takes 10ms computing the additional layers.
Our region-wise computation is also lower, thanks to fewer proposals (300 per image). Our system has a frame-rate of 17 fps with the ZF net.

\setlength{\tabcolsep}{2pt}
\renewcommand{\arraystretch}{1.1}
\begin{table}[t]
\begin{center}
\caption{Detection results of Faster R-CNN on PASCAL VOC 2007 test set using \textbf{different settings of anchors}. The network is VGG-16. The training data is VOC 2007 trainval. The default setting of using 3 scales and 3 aspect ratios (69.9\%) is the same as that in Table~\ref{tab:vgg}.}
\vspace{-1em}
\small
\begin{tabular}{c|c|c|c}
settings &  anchor scales & aspect ratios & mAP (\%)\\
\hline
\hline
\multirow{2}{*}{1 scale, 1 ratio} & \footnotesize $128^2$ & 1:1 & 65.8 \\
                                  & \footnotesize $256^2$ & 1:1 & 66.7 \\
\hline
\multirow{2}{*}{1 scale, 3 ratios} & \footnotesize $128^2$ & \{2:1, 1:1, 1:2\} & 68.8 \\
                                   & \footnotesize $256^2$ & \{2:1, 1:1, 1:2\} & 67.9 \\
\hline
\multirow{1}{*}{3 scales, 1 ratio} & \footnotesize $\{128^2, 256^2, 512^2\}$ & 1:1 & \textbf{69.8} \\
\hline
\multirow{1}{*}{3 scales, 3 ratios} &  \footnotesize $\{128^2, 256^2, 512^2\}$ & \{2:1, 1:1, 1:2\} & \textbf{69.9} \\
\end{tabular}
\label{tab:anchors}
\end{center}
\end{table}

\setlength{\tabcolsep}{10pt}
\begin{table}[t]
\begin{center}
\caption{Detection results of Faster R-CNN on PASCAL VOC 2007 test set using \textbf{different values of $\lambda$} in Equation~(\ref{eq:loss}). The network is VGG-16. The training data is VOC 2007 trainval. The default setting of using $\lambda=10$ (69.9\%) is the same as that in Table~\ref{tab:vgg}.}
\vspace{-1em}
\small
\begin{tabular}{c|cccc}
$\lambda$ & 0.1 & 1 & 10 & 100\\
\hline
mAP (\%) & 67.2 & 68.9 & 69.9 & 69.1\\
\end{tabular}
\label{tab:lambda}
\end{center}
\end{table}

\begin{figure*}[t]
\centering
\includegraphics[width=0.7\linewidth]{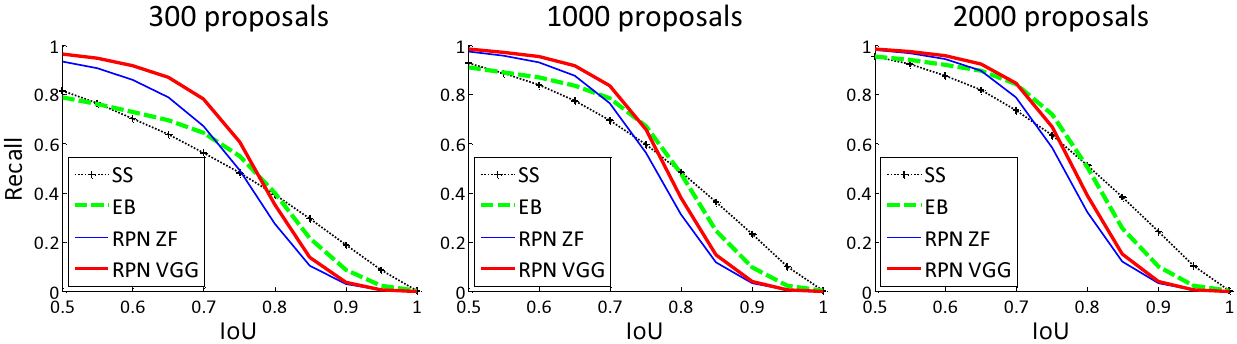}
\caption{Recall \vs~IoU overlap ratio on the PASCAL VOC 2007 test set.}
\label{fig:recall}
\end{figure*}

\setlength{\tabcolsep}{6pt}
\renewcommand{\arraystretch}{1.1}
\begin{table*}[t]
\begin{center}
\caption{\textbf{One-Stage Detection \vs~Two-Stage Proposal + Detection}. Detection results are on the PASCAL VOC 2007 test set using the ZF model and Fast R-CNN. RPN uses unshared features.}
\vspace{-1em}
\small
\begin{tabular}{c|cc|c|c}
   & \multicolumn{2}{c|}{proposals} & detector & mAP (\%) \\
  \hline\hline
  Two-Stage & RPN + ZF, unshared & 300 & Fast R-CNN + ZF, 1 scale & 58.7 \\
    \hline
  One-Stage & dense, 3 scales, 3 aspect ratios & 20000 & Fast R-CNN + ZF, 1 scale & 53.8 \\
  One-Stage & dense, 3 scales, 3 aspect ratios & 20000 & Fast R-CNN + ZF, 5 scales & 53.9 \\
\end{tabular}
\label{tab:overfeat}
\end{center}
\end{table*}

\vspace{.5em}
\noindent\textbf{Sensitivities to Hyper-parameters.} In Table~\ref{tab:anchors} we investigate the settings of anchors. By default we use 3 scales and 3 aspect ratios (69.9\% mAP in Table~\ref{tab:anchors}).
If using just one anchor at each position, the mAP drops by a considerable margin of 3-4\%. The mAP is higher if using 3 scales (with 1 aspect ratio) or 3 aspect ratios (with 1 scale), demonstrating that using anchors of multiple sizes as the regression references is an effective solution. Using just 3 scales with 1 aspect ratio (69.8\%) is as good as using 3 scales with 3 aspect ratios on this dataset, suggesting that scales and aspect ratios are not disentangled dimensions for the detection accuracy. But we still adopt these two dimensions in our designs to keep our system flexible.

In Table~\ref{tab:lambda} we compare different values of $\lambda$ in Equation~(\ref{eq:loss}). By default we use $\lambda=10$ which makes the two terms in Equation~(\ref{eq:loss}) roughly equally weighted after normalization. Table~\ref{tab:lambda} shows that our result is impacted just marginally (by $\sim1\%$) when $\lambda$ is within a scale of about two orders of magnitude (1 to 100). This demonstrates that the result is insensitive to $\lambda$ in a wide range.

\vspace{.5em}
\noindent\textbf{Analysis of Recall-to-IoU.}
Next we compute the recall of proposals at different IoU ratios with ground-truth boxes. It is noteworthy that the Recall-to-IoU metric is just \emph{loosely} \cite{Hosang2014,Hosang2015,Chavali2015} related to the ultimate detection accuracy. It is more appropriate to use this metric to \emph{diagnose} the proposal method than to evaluate it.

In Figure~\ref{fig:recall}, we show the results of using 300, 1000, and 2000 proposals. We compare with SS and EB, and the $N$ proposals are the top-$N$ ranked ones based on the confidence generated by these methods.
The plots show that the RPN method behaves gracefully when the number of proposals drops from 2000 to 300. This explains why the RPN has a good ultimate detection mAP when using as few as 300 proposals. As we analyzed before, this property is mainly attributed to the \emph{cls} term of the RPN. The recall of SS and EB drops more quickly than RPN when the proposals are fewer.

\vspace{.5em}
\noindent\textbf{One-Stage Detection \vs~Two-Stage Proposal + Detection.}
The OverFeat paper \cite{Sermanet2014} proposes a detection method that uses regressors and classifiers on sliding windows over convolutional feature maps. OverFeat is a \emph{one-stage}, \emph{class-specific} detection pipeline, and ours is a \emph{two-stage cascade} consisting of class-agnostic proposals and class-specific detections. In OverFeat, the region-wise features come from a sliding window of one aspect ratio over a scale pyramid. These features are used to simultaneously determine the location and category of objects. In RPN, the features are from square (3$\times$3) sliding windows and predict proposals relative to anchors with different scales and aspect ratios. Though both methods use sliding windows, the region proposal task is only the first stage of Faster R-CNN---the downstream Fast R-CNN detector \emph{attends} to the proposals to refine them. In the second stage of our cascade, the region-wise features are adaptively pooled \cite{He2014,Girshick2015a} from proposal boxes that more faithfully cover the features of the regions. We believe these features lead to more accurate detections.

To compare the one-stage and two-stage systems, we \emph{emulate} the OverFeat system (and thus also circumvent other differences of implementation details) by \emph{one-stage} Fast R-CNN. In this system, the ``proposals'' are dense sliding windows of 3 scales (128, 256, 512) and 3 aspect ratios (1:1, 1:2, 2:1).
Fast R-CNN is trained to predict class-specific scores and regress box locations from these sliding windows. Because the OverFeat system adopts an image pyramid, we also evaluate using convolutional features extracted from 5 scales. We use those 5 scales as in \cite{He2014,Girshick2015a}.

Table~\ref{tab:overfeat} compares the two-stage system and two variants of the one-stage system. Using the ZF model, the one-stage system has an mAP of 53.9\%. This is lower than the two-stage system (58.7\%) by 4.8\%. This experiment justifies the effectiveness of cascaded region proposals and object detection. Similar observations are reported in \cite{Girshick2015a,Lenc2015}, where replacing SS region proposals with sliding windows leads to $\sim$6\% degradation in both papers.
We also note that the one-stage system is slower as it has considerably more proposals to process.

\setlength{\tabcolsep}{2pt}
\renewcommand{\arraystretch}{1.1}
\begin{table*}[t]
\begin{center}
\caption{Object detection results (\%) on the \textbf{MS COCO} dataset. The model is VGG-16.}
\vspace{-1em}
\small
\begin{tabular}{l|c|c|c|c|c|c}
       &           &               & \multicolumn{2}{c|}{COCO val} & \multicolumn{2}{c}{COCO test-dev}\\\cline{4-7}
method & proposals & training data & ~~~mAP@.5~~~ & mAP@[.5, .95]  & ~~~mAP@.5~~~ & mAP@[.5, .95] \\
\hline\hline
Fast R-CNN \cite{Girshick2015a} & SS, 2000 & COCO train & - & - & 35.9 & 19.7 \\
Fast R-CNN \footnotesize [impl. in this paper] & SS, 2000 & COCO train & 38.6 & 18.9 & 39.3 & 19.3\\
\hline
Faster R-CNN & RPN, 300 & COCO train & 41.5 & 21.2 & 42.1 & 21.5 \\
Faster R-CNN & RPN, 300 & COCO trainval & - & - & \textbf{42.7} & \textbf{21.9} \\
\end{tabular}
\label{tab:coco}
\end{center}
\end{table*}

\subsection{Experiments on MS COCO}

We present more results on the Microsoft COCO object detection dataset \cite{Lin2014}. This dataset involves 80 object categories. We experiment with the 80k images on the training set, 40k images on the validation set, and 20k images on the test-dev set.
We evaluate the mAP averaged for IoU $\in[0.5:0.05:0.95]$ (COCO's standard metric, simply denoted as mAP@[.5, .95]) and mAP@0.5 (PASCAL VOC's metric).

There are a few minor changes of our system made for this dataset. We train our models on an 8-GPU implementation, and the effective mini-batch size becomes 8 for RPN (1 per GPU) and 16 for Fast R-CNN (2 per GPU). The RPN step and Fast R-CNN step are both trained for 240k iterations with a learning rate of 0.003 and then for 80k iterations with 0.0003. We modify the learning rates (starting with 0.003 instead of 0.001) because the mini-batch size is changed.
For the anchors, we use 3 aspect ratios and 4 scales (adding $64^2$), mainly motivated by handling small objects on this dataset. In addition, in our Fast R-CNN step, the negative samples are defined as those with a maximum IoU with ground truth in the interval of $[0, 0.5)$, instead of $[0.1, 0.5)$ used in \cite{He2014,Girshick2015a}. We note that in the SPPnet system \cite{He2014}, the negative samples in $[0.1, 0.5)$ are used for network fine-tuning, but the negative samples in $[0, 0.5)$ are still visited in the SVM step with hard-negative mining. But the Fast R-CNN system \cite{Girshick2015a} abandons the SVM step, so the negative samples in $[0, 0.1)$ are never visited. Including these $[0, 0.1)$ samples improves mAP@0.5 on the COCO dataset for both Fast R-CNN and Faster R-CNN systems (but the impact is negligible on PASCAL VOC).

The rest of the implementation details are the same as on PASCAL VOC. In particular, we keep using 300 proposals and single-scale ($s=600$) testing. The testing time is still about 200ms per image on the COCO dataset.

In Table~\ref{tab:coco} we first report the results of the Fast R-CNN system \cite{Girshick2015a} using the implementation in this paper. Our Fast R-CNN baseline has 39.3\% mAP@0.5 on the test-dev set, higher than that reported in \cite{Girshick2015a}. We conjecture that the reason for this gap is mainly due to the definition of the negative samples and also the changes of the mini-batch sizes. We also note that the mAP@[.5, .95] is just comparable.

\begin{figure*}[t]
\centering
\includegraphics[width=0.72\linewidth]{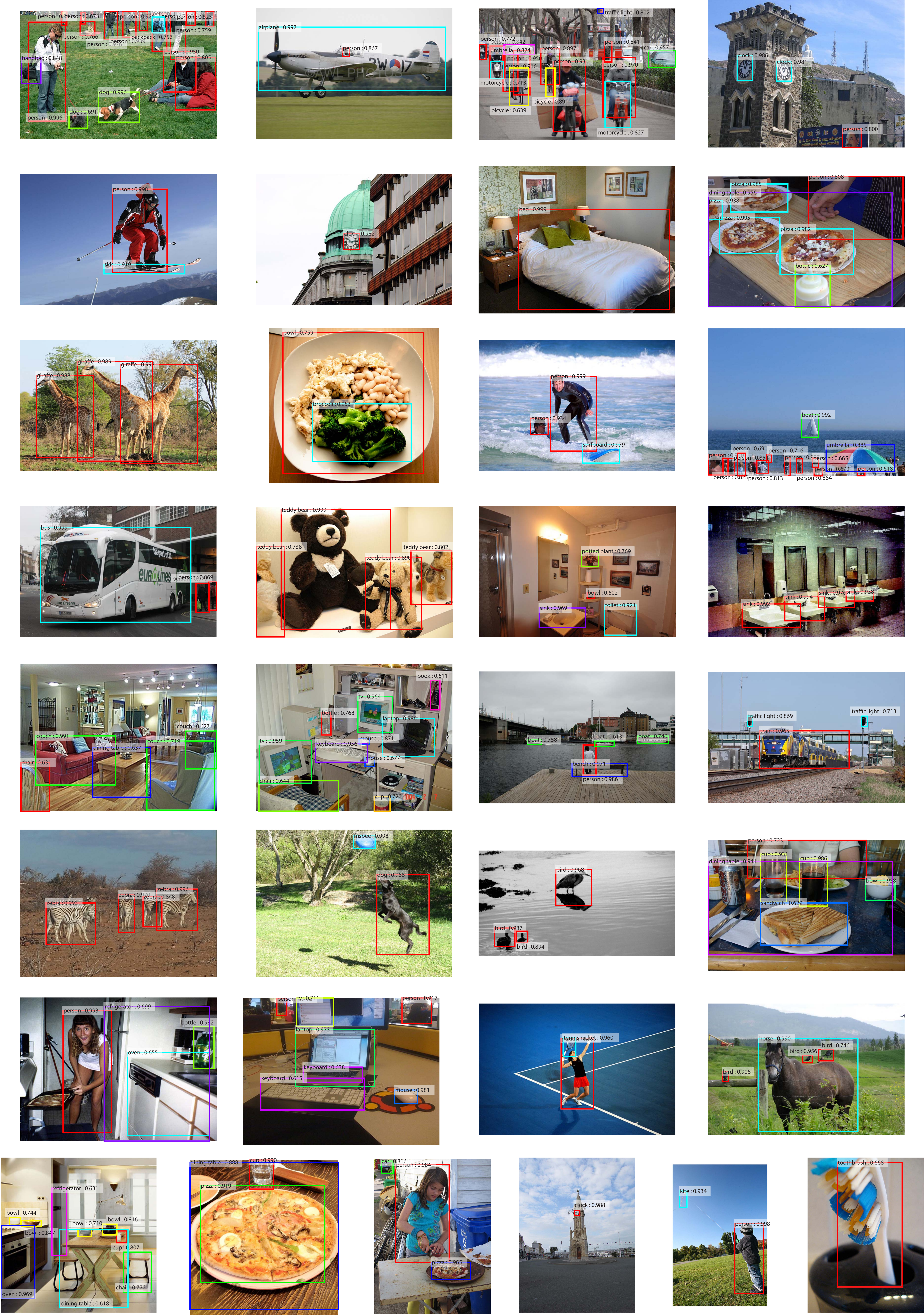}
\caption{Selected examples of object detection results on the MS COCO test-dev set using the Faster R-CNN system. The model is VGG-16 and the training data is COCO trainval (42.7\% mAP@0.5 on the test-dev set). Each output box is associated with a category label and a softmax score in $[0,1]$. A score threshold of 0.6 is used to display these images. For each image, one color represents one object category in that image.}
\label{fig:coco_results}
\end{figure*}

Next we evaluate our Faster R-CNN system.
Using the COCO training set to train, Faster R-CNN has 42.1\% mAP@0.5 and 21.5\% mAP@[.5, .95] on the COCO test-dev set. This is 2.8\% higher for mAP@0.5 and \textbf{2.2\% higher for mAP@[.5, .95]} than the Fast R-CNN counterpart under the same protocol (Table~\ref{tab:coco}). This indicates that RPN performs excellent for improving the localization accuracy at higher IoU thresholds.
Using the COCO trainval set to train, Faster R-CNN has 42.7\% mAP@0.5 and 21.9\% mAP@[.5, .95] on the COCO test-dev set.
Figure~\ref{fig:coco_results} shows some results on the MS COCO test-dev set.

\vspace{.5em}
\noindent\textbf{Faster R-CNN in ILSVRC \& COCO 2015 competitions}
We have demonstrated that Faster R-CNN benefits more from better features, thanks to the fact that the RPN completely learns to propose regions by neural networks. This observation is still valid even when one increases the depth substantially to over 100 layers \cite{He2015a}. Only by replacing VGG-16 with a 101-layer residual net (ResNet-101) \cite{He2015a}, the Faster R-CNN system increases the mAP from 41.5\%/21.2\% (VGG-16) to 48.4\%/27.2\% (ResNet-101) on the COCO val set. With other improvements orthogonal to Faster R-CNN, He \etal \cite{He2015a} obtained a single-model result of 55.7\%/34.9\% and an ensemble result of 59.0\%/37.4\% on the COCO test-dev set, which won the 1st place in the COCO 2015 object detection competition. The same system \cite{He2015a} also won the 1st place in the ILSVRC 2015 object detection competition, surpassing the second place by absolute 8.5\%. RPN is also a building block of the 1st-place winning entries in ILSVRC 2015 localization and COCO 2015 segmentation competitions, for which the details are available in \cite{He2015a} and \cite{Dai2015a} respectively.

\subsection{From MS COCO to PASCAL VOC}

\setlength{\tabcolsep}{8pt}
\renewcommand{\arraystretch}{1.1}
\begin{table}[t]
\begin{center}
\caption{Detection mAP (\%) of Faster R-CNN on PASCAL VOC 2007 test set and 2012 test set using different training data. The model is VGG-16. ``COCO'' denotes that the COCO trainval set is used for training. See also Table~\ref{tab:vgg_all} and Table~\ref{tab:vgg12_all}.}
\vspace{-1em}
\small
\begin{tabular}{l|c|c}
  training data & 2007 test & 2012 test \\
  \hline\hline
  VOC07 & 69.9 & 67.0 \\
  VOC07+12 & 73.2 & - \\
  VOC07++12 & - & 70.4 \\
  \hline
  COCO (no VOC) & 76.1 & 73.0 \\
  COCO+VOC07+12 & \textbf{78.8} & - \\
  COCO+VOC07++12 & - & \textbf{75.9} \\
\end{tabular}
\label{tab:voc_coco}
\end{center}
\end{table}

Large-scale data is of crucial importance for improving deep neural networks. Next, we investigate how the MS COCO dataset can help with the detection performance on PASCAL VOC.

As a simple baseline, we directly evaluate the COCO detection model on the PASCAL VOC dataset, \emph{without fine-tuning on any PASCAL VOC data}. This evaluation is possible because the categories on COCO are a superset of those on PASCAL VOC. The categories that are exclusive on COCO are ignored in this experiment, and the softmax layer is performed only on the 20 categories plus background.
The mAP under this setting is 76.1\% on the PASCAL VOC 2007 test set (Table~\ref{tab:voc_coco}). This result is better than that trained on VOC07+12 (73.2\%) by a good margin, even though the PASCAL VOC data are not exploited.

Then we fine-tune the COCO detection model on the VOC dataset. In this experiment, the COCO model is in place of the ImageNet-pre-trained model (that is used to initialize the network weights), and the Faster R-CNN system is fine-tuned as described in Section~\ref{sec:training}. Doing so leads to 78.8\% mAP on the PASCAL VOC 2007 test set. The extra data from the COCO set increases the mAP by 5.6\%. Table~\ref{tab:vgg_all} shows that the model trained on COCO+VOC has the best AP for every individual category on PASCAL VOC 2007.
Similar improvements are observed on the PASCAL VOC 2012 test set (Table~\ref{tab:voc_coco} and Table~\ref{tab:vgg12_all}). We note that the test-time speed of obtaining these strong results is still about \textbf{200ms per image}.

\section{Conclusion}

We have presented RPNs for efficient and accurate region proposal generation. By sharing convolutional features with the down-stream detection network, the region proposal step is nearly cost-free. Our method enables a unified, deep-learning-based object detection system to run at near real-time frame rates. The learned RPN also improves region proposal quality and thus the overall object detection accuracy.

\bibliographystyle{IEEEtran}
\bibliography{IEEEabrv,rpn_pami_arxiv}

\end{document}